\documentclass[letterpaper, 10 pt, conference]{ieeeconf}  

\IEEEoverridecommandlockouts                              

\usepackage{times} 

\usepackage{float}
\usepackage{newtxmath}        
\usepackage{graphicx}
\usepackage{hyperref}

\title{\LARGE \bf
MARTIAN: A Rendering Framework for Aerial Mars Imagery from HiRISE Orbital Data
}

\author{
Dario Pisanti$^{1,*}$,
Georgios Georgakis$^{2}$%
\thanks{$^{1}$Space Robotics Research Group, SnT, University of Luxembourg.}%
\thanks{$^{2}$Jet Propulsion Laboratory, California Institute of Technology.}%
\thanks{$^{*}$Corresponding author: dario.pisanti@uni.lu.}%
}

\begin{document}

\maketitle
\thispagestyle{empty}
\pagestyle{empty}

\begin{abstract}

Aerial navigation on Mars requires vision-based pipelines that are robust to the diverse illumination conditions and terrain morphology of the Martian surface. A key bottleneck for training and evaluating such methods is the scarcity of large-scale, annotated aerial datasets. We present MARTIAN, an open-source Blender-based rendering framework that leverages real HiRISE orbital map products to synthesize realistic aerial views of the Martian terrain under controllable lighting conditions and at varying altitudes. MARTIAN generates observations with accurate pose annotations, directly addressing the scarcity of training data for vision-based navigation on Mars. The framework has been validated through its deployment in concurrent work on map-based localization systems for Ingenuity and future Mars rotorcraft, where synthetically trained deep image matchers were successfully evaluated on real Mars imagery. MARTIAN is publicly available at: \href{https://github.com/nasa-jpl/martian}{https://github.com/nasa-jpl/martian}.

\end{abstract}


\section{INTRODUCTION}
\label{sec:intro}

The demonstration flights of NASA's Mars Helicopter, \textit{Ingenuity}, have marked a groundbreaking milestone in Mars exploration~\cite{Grip2022}. One of the next evolutionary steps is the Mars Science Helicopter (MSH)~\cite{Tzanetos2022}, a hexacopter concept with multi-kg payload capacity and 10 km traverse capability at altitudes up to 100 m, aimed at supporting high-priority investigations in Martian astrobiology, climate, and geology~\cite{Bapst2021Mars}. Similarly, the Skyfall mission aims to deploy six scout helicopters mid-air in the Martian atmosphere from the entry capsule, thus eliminating the need for a landing platform~\cite{skyfall2026nasa}. The deployed UAVs will independently perform exploration tasks for resource prospecting, science measurements, and safe landing site inspection for future crewed missions.

These aerial platforms will need to perform precision onboard vision-based navigation and localization to enable long-range flights over uncertain terrain in a global navigation satellite system (GNSS)-denied environment. Images captured by the onboard navigation camera are typically processed by a Visual Inertial Odometry (VIO) system to produce position estimates in a relative navigation fashion. However, VIO drift needs to be bounded over long traverses with an absolute localization strategy. This can be accomplished by registering onboard imagery onto orbital maps pre-registered to a global reference frame and estimating the navigation camera pose relative to the map. This inherently drift-free geo-localization technique is referred to as Map-based Localization (MbL).

Both VIO and MbL methods rely on the identification of distinctive features in onboard imagery and orbital maps to enable relative and absolute navigation, respectively. This process poses significant perception challenges due diverse illumination conditions and terrain morphology, and is even more pronounced in MbL due to large differences in lighting and scale between onboard imagery and the reference map. This problem is also relevant for vision-based Terrain Relative Navigation during the Entry, Descent and Landing (EDL) phase of Martian landers\cite{johnson2023}, which rely on orbital maps taken at restricted times of day.

While recent deep learning methods~\cite{loftr,roma} have demonstrated robustness to illumination and scale variations on in-the-wild datasets such as MegaDepth~\cite{megadepth} for terrestrial applications, the main bottleneck is the lack of large-scale datasets that would allow finetuning these methods in planetary domains. Unlike Earth-based applications relying on abundant data~\cite{zhu2023sues}, large-scale and annotated aerial datasets are not readily available for Mars. Imagery data from the Mars2020 mission provide aerial observations from the Landing Vision System (LVS) camera during EDL and from the navigation camera onboard Ingenuity. However, the lack of accurate pose annotations prevents these data to be used for a comprehensive robustness study on scale and illumination variations.
Existing large-scale Martian datasets, such as AI4MARS~\cite{swan2021ai4mars}, address surface-based navigation by providing terrain classification labels for rover driving, while the SynMars-Air dataset~\cite{synmars-air} addresses aerial object detection and terrain segmentation from a helicopter viewpoint.

To better support learning-based perception methods for aerial navigation on Mars, we developed the Mars Aerial Rendering Tool for Imaging and Navigation (MARTIAN). This simulation framework leverages real map products from the Mars Reconnaissance Orbiter (MRO) High-Resolution Imaging Science Experiment (HiRISE)~\cite{hirise} to generate large-scale datasets for training vision-based navigation pipelines. MARTIAN has been deployed in concurrent work, as detailed in Section \ref{sec:applications}.

\section{MARTIAN framework}
\label{sec:martian}

\begin{figure}
    \centering
    \includegraphics[width=\linewidth]{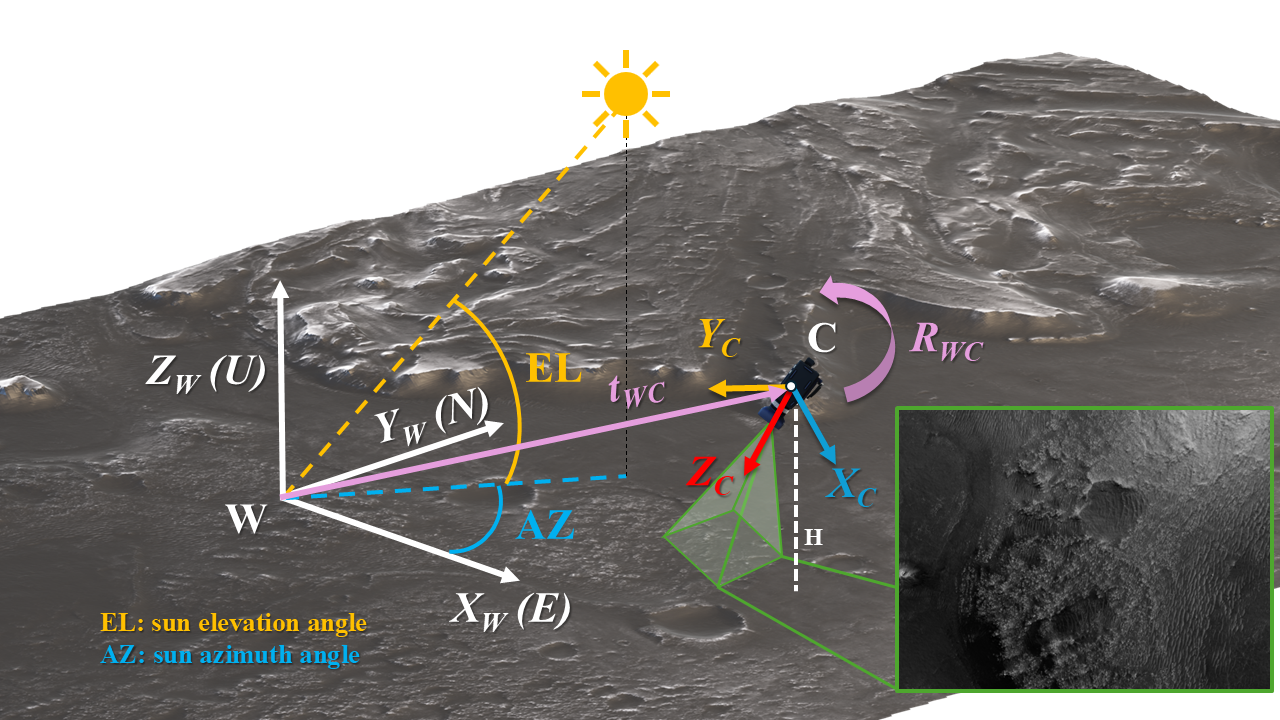}  \caption{\label{fig:MARTIAN_view}View of the Jezero crater site in MARTIAN with the simulated perspective imaging camera and a rendered observation.}
\end{figure}
\begin{figure}
    \centering
    \includegraphics[width=\linewidth]{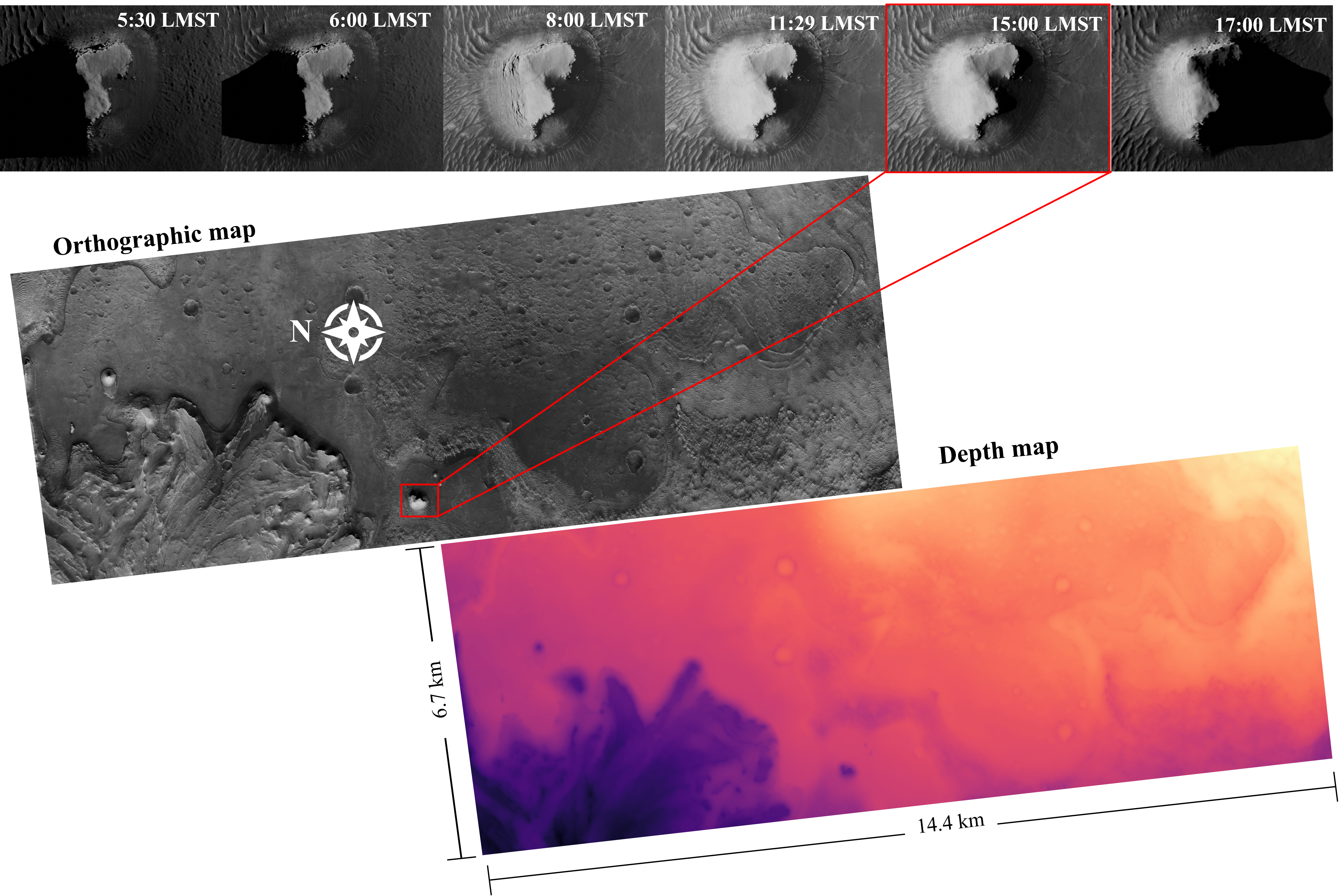}  \caption{\label{fig:renderings}Ortographic grayscale map rendered in MARTIAN at 15:00 Local Mean Solar Time (LMST) on the Jezero crater site on 2031-05-10, with the corresponding depth map. Simulated aerial observations from the same Martian day are shown from 05:30 to 17:00 LMST.  }
\end{figure}

We take advantage of real map products created from HiRISE to create a large-scale dataset suitable for training vision-based pipelines for aerial navigation on Mars. We developed a python-based framework in the open-source 3D computer graphics software Blender to import HiRISE Digital Terrain Models (DTMs) and textured ortho-projected images to generate maps and aerial observations of a Martian site under various lighting conditions and at different altitudes.
\\
\\
\noindent \textbf{HiRISE data.}
With its resolution capability at nadir of 25 centimeters per pixel from 300 km altitude, HiRISE has been serving as an indispensable orbital asset for identifying and selecting safe landing sites for robotics exploration missions. 
In this work, we utilize a 1 m / post DTM and a 0.25 m / pixel ortho-image with equirectangular projection generated from stereo pairs imaging of the Jezero Crater landing site for the Mars2020 Mission, over an area of 6.737 km by 14.403 km. 
\\
\\
\noindent \textbf{Terrain and texture modeling.}
The Jezero HiRISE DTM was imported in Blender 4.0 using a modified version of the original HiRISE import plug-in \cite{BlenderHiRISEDTMImporter}. This add-on can load the terrain data at a desired resolution within a range (0., 100] $\%$ of the full model resolution and generate a mesh. By leveraging terrain metadata, including geographic information and resolution, a UV map is created for ortho-images to be draped over the companion DTM and used as a high-fidelity terrain texture. 
The MARTIAN framework initially imports the Jezero DTM with 10$\%$ of its original resolution, allowing for efficient terrain setup. Then, a material is added to the mesh, with its surface shading model being controlled via a Principled Bidirectional Scattering Distribution Function (BSDF).
The 0.25 m / pixel Jezero ortho-image is then loaded in the shading editor as the base texture for the terrain surface. The ortho-image coordinates data are retrieved by a Texture Coordinate Node to ensure that the texture is properly mapped onto the 3D mesh. Finally, the mesh is reloaded with its full resolution of 1 m / post.
\\
\\
\noindent \textbf{Scene setup and camera modeling.}
MARTIAN allows for setting multiple scenes for perspective and orthographic imaging with user-defined camera intrinsics and extrinsics. Figure~\ref{fig:MARTIAN_view} shows a view of the Jezero crater site in MARTIAN. Given the \textit{world} reference frame $W$ defined as an East-North-Up coordinate system with origin on the terrain map center, the camera can be located in $W$ by providing the xy-coordinates and the altitude (in meters) with respect to the terrain at those coordinates. To position the camera object above the terrain mesh at the desired altitude, MARTIAN adopts a ray-tracing approach by using a Bounding Volume Hierarchy tree, a data structure used by Blender to efficiently organize geometric objects in 3D space. The camera frame $C$ is centered at the optical center, with its X-axis pointing to right along the image width, the Z-axis pointing towards the terrain, and the Y-axis completing the orthogonal set. The attitude of the $C$  with respect to the world frame $W$ is specified as input, and the final pose ($\mathbf{R}_{WC} \vert \mathbf{t}_{WC}$) is saved, where $\mathbf{t}_{WC}$ is the location of the camera in world coordinated and $\mathbf{R}_{WC}$ is the rotation matrix that aligns $W$ to $C$.
\\
\\
\noindent \textbf{Lighting.}
MARTIAN provides the capability to render scenes in Blender with various illumination conditions by tuning Sun light source parameters such as irradiance, apparent disk diameter and orientation in the map frame. Higher irradiance values cast brighter illumination and shadows in the scene, while the angular diameter of the Sun disk as seen by the scene controls the softness versus harshness of the shadows. Sun orientation simulating a specific time of day is specified by the user through elevation (EL) and azimuth angles (AZ) as shown in Figure \ref{fig:MARTIAN_view}. Lighting computations are performed using the Blender Cycles engine. This is a physically-based rendering engine that uses a ray tracing algorithm to accurately simulate light behavior. 
Sun irradiance is set to the maximum value of 590 W/m$^2$ and angular diameter to 0.35°, consistent with actual estimations for Mars.The BSDF parameters and camera exposure were tuned empirically to match the visual appearance of the real ortho-projected map. Figure \ref{fig:renderings} illustrates the effect of different times of day on Mars on a rendered aerial observation from the Jezero crater site.

\section{Applications in Aerial Map-based Localization}
\label{sec:applications}
MARTIAN was used to generate a large-scale synthetic aerial dataset to train deep image matchers for the autonomous Map-Based Localization (MbL) of Mars rotorcraft. The dataset comprises 17 orthographic grayscale maps rendered from combinations of Sun azimuth ($0^{\circ}-360^{\circ}$, $45^{\circ}$ steps) and elevation ($30^{\circ}$, $60^{\circ}$, $90^{\circ}$), one corresponding orthographic depth map, and 4500 nadir-pointing aerial observations at fixed Sun angles AZ=180$^{\circ}$ and EL=40$^{\circ}$ with their corresponding depth images. Observations were randomly sampled from the HiRISE Jezero Crater DTM with uniform distribution in the $(x,y)$ world-frame coordinates and within the altitude range [$64, 200$] m., where the 64 m lower bound reflects MARTIAN's resolution limit, tied to the HiRISE texture resolution. Camera extrinsics and altitude data were stored for each observation and used as ground truth.
\\

\noindent\textbf{MbL for Ingenuity \cite{georgakis2025}.} This dataset was used to pre-train LoFTR~\cite{loftr}, a detector-free image matching method that uses Transformers~\cite{transformer} in a coarse-to-fine strategy to produce pixel-wise semi-dense correspondences. The pre-trained model was then incorporated into an MbL system for Ingenuity, with a training strategy designed to address the scarcity of annotated in-domain data. Since only a limited number of Ingenuity navigation images are available and obtaining accurate pose annotations is particularly challenging due to noisy VIO estimates and frequent low-texture terrain, the approach first leverages MARTIAN data to obtain a suitable initial representation for HiRISE map features, before fine-tuning on a small set of Ingenuity images with self-supervised pseudo annotations. This intermediate pre-training on rendered data proved beneficial: fine-tuning LoFTR directly on Ingenuity images without it resulted in an 11.2\% Acc@5m drop across all flights, and a gap of up to 26.4\% on the most challenging terrain. The MbL pipeline achieved 89.4\% Acc@5m and near-perfect 99.8\% Acc@10m across Ingenuity flights, outperforming prior template-matching and hand-crafted feature methods, while generalizing to previously unseen flights with as few as 30 training images from a single flight.
\\

\begin{figure}
    \centering
    \includegraphics[width=\linewidth]{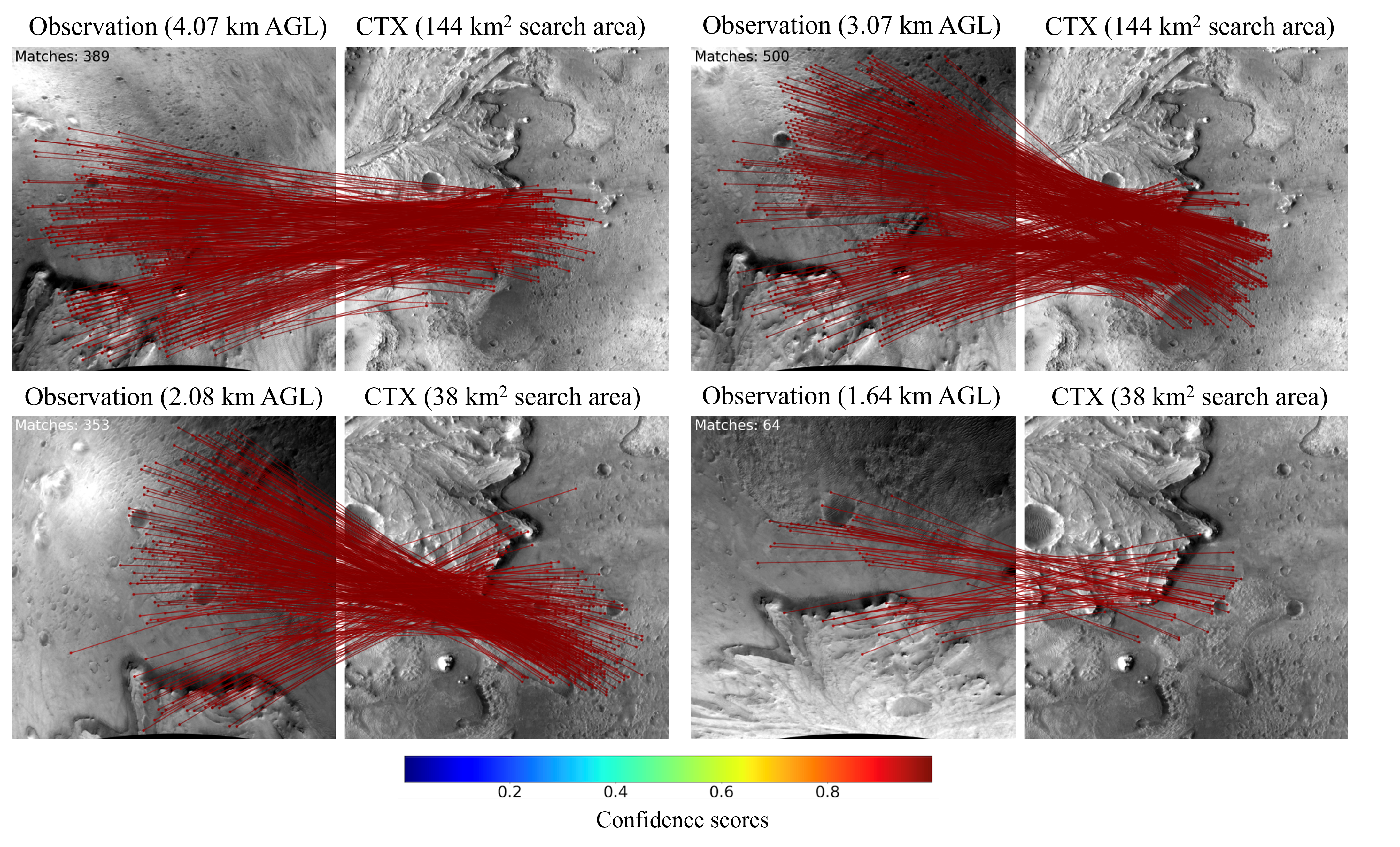}  \caption{\label{fig:mars2020}Geo-LoFTR matches between CTX map crops and LCAM observations at different altitudes Above Ground Level (AGL) from the imaging sequence during the Mars2020 EDL.}
\end{figure}

\noindent\textbf{MbL for future Mars rotorcraft under challenging lighting \cite{pisanti2025vision}.} The same dataset was used to train Geo-LoFTR, a geometry-aided extension of LoFTR that incorporates geometric context from DTMs derived from orbital data products into the feature matching process. Geo-LoFTR was adopted in an MbL pipeline targeting the long-range navigation of future Mars rotorcraft at altitudes up to 200 m, such as the Mars Science Helicopter (MSH)\cite{Tzanetos2022}, demonstrating improved localization accuracy over prior methods by up to 31.8\% Acc@1m under challenging illumination conditions. In particular, the geometric context was shown to increase robustness to varying Sun elevation and azimuth angles, as well as to the scale difference between onboard observations and the reference map, where Geo-LoFTR maintained consistent performance across the full altitude range. Robustness was also demonstrated over a full simulated Martian day at the Jezero crater site, from 6:00 to 17:00 Local Mean Solar Time (LMST), against a HiRISE-like reference map captured at 15:00 LMST.
Furthermore, Geo-LoFTR was validated on real Mars2020 descent imagery, where frames captured by the onboard Lander Vision System camera (LCAM)\cite{johnson2023} were matched against a Context Camera (CTX) map~\cite{malin2007context} at 6 m/px resolution. Despite being trained entirely on MARTIAN-rendered data for the anticipated flight altitude of MSH, Geo-LoFTR successfully provided global localization solutions across descent frames in the 6 km -960 m altitude range, indicating preliminary synthetic-to-real transfer under comparable observation-to-map scale ratios. Figure~\ref{fig:mars2020} shows an example of Geo-LoFTR matches between LCAM frame captured at different altitudes and CTX map crops obtained by leveraging pose priors. However, broader generalization across terrain types and map sources still needs further investigation.

\section{Conclusion}
\label{sec:workshop_conclusion}

We presented MARTIAN, a Blender-based rendering framework that leverages HiRISE map products to generate large-scale annotated aerial datasets for Mars, addressing a critical gap for training vision-based navigation pipelines. MARTIAN has already demonstrated utility in concurrent work on map-based localization for both Ingenuity and future Mars rotorcraft.
A current limitation is that HiRISE textures are ortho-images captured at a specific time of day, and therefore already embed the illumination and shadow conditions of that acquisition, which the renderer cannot fully decouple. Additionally, while preliminary validation of MARTIAN-trained methods on real Mars imagery suggests a promising sim-to-real transfer, further studies across diverse landing sites are needed to fully characterize generalization.






\bibliographystyle{IEEEtran}
\bibliography{IEEEabrv, references}

\end{document}